\documentclass[letterpaper, 10 pt, conference]{ieeeconf} 

\IEEEoverridecommandlockouts                              

\usepackage{graphicx}
\usepackage{subcaption}
\usepackage{epsfig}
\usepackage{mathptmx}
\usepackage{times}
\usepackage{amsmath} 
\usepackage{amssymb}
\usepackage{siunitx}

\title{\LARGE \bf
Neural Network-Based In-Contact Task Segmentation \\ of Demonstrated Robot Motions via Hand Guiding
}

\author{Johannes Hartwig$^{1}$, Marcello Kögler$^{1}$ and Dominik Henrich$^{1}$
\thanks{$^{1}$  Chair for Applied Computer Science III (Robotics and Embedded Systems),
                University of Bayreuth, 95440 Bayreuth, Germany,
                {\tt\footnotesize johannes.hartwig@uni-bayreuth.de}}%
}

\begin{document}

\maketitle
\thispagestyle{empty}
\pagestyle{empty}

\begin{abstract}
Intuitive robot programming approaches, such as Programming by Demonstration, enable non-experts to teach robots tasks through hand-guided demonstrations. 
However, for in-contact tasks requiring force control for parts of the motion, automated segmentation of motion types is crucial for improving usability. 
Considering the related work, this work proposes a neural network-based segmentation method categorizing motion into three fundamental types using input data of cartesian twists and wrenches at the end effector. 
We collect demonstration data using a 7-axis robot with a wrist-mounted force torque sensor and generate ground truth labeling by hand.
While still in development, our research direction aims to enable non-experts to program in-contact tasks more intuitively.
\end{abstract}

\section{Introduction}

The shortage of skilled workers significantly affects the economy, making automation a key strategy to enhance work efficiency \cite{Zwysen2023}.
As more affordable robot models enter the market \cite{Adachi2024}, cobots are also becoming accessible to small and medium-sized enterprises (SMEs).
Nevertheless, hiring a robot expert to program such systems is often not affordable for SMEs. 
Intuitive robot programming approaches focus on simplifying robotic systems and minimizing this problem by enabling non-experts to program them \cite{Colceriu2020}.

Programming by Demonstration (PbD) provides an intuitive solution by allowing users, for example, to move the robot along the desired path via hand guiding \cite{Orendt2016, Villani2018}.
However, the robot's trajectory alone is sometimes not sufficient to achieve all the goals intended by the user. 
For some tasks, the interaction force with the environment is also required (e.g., during surface treatment or assembly). 
Therefore, these so-called in-contact or contact-rich tasks require a specification of position and force over time. 
To enable non-experts to program such tasks, we extend an existing intuitive programming framework \cite{Riedl2019, Riedl2021} for pick-and-place tasks to in-contact tasks.
For this, we developed an intuitive visualization and force input method via a touch GUI in our previous work \cite{Hartwig2023a, Hartwig2023b}. 
To further extend our approach, measuring the interaction forces with the environment directly during the user's demonstration would be beneficial.
Often, a wrist-mounted force torque sensor \cite{Steinmetz2015, Conkey2019} or a sensor attached to the environment \cite{Bargmann2021} is utilized.

\begin{figure*}[t!]
    \centering
    \begin{subfigure}[t]{0.32\textwidth}
        \centering
        \includegraphics[width=\textwidth]{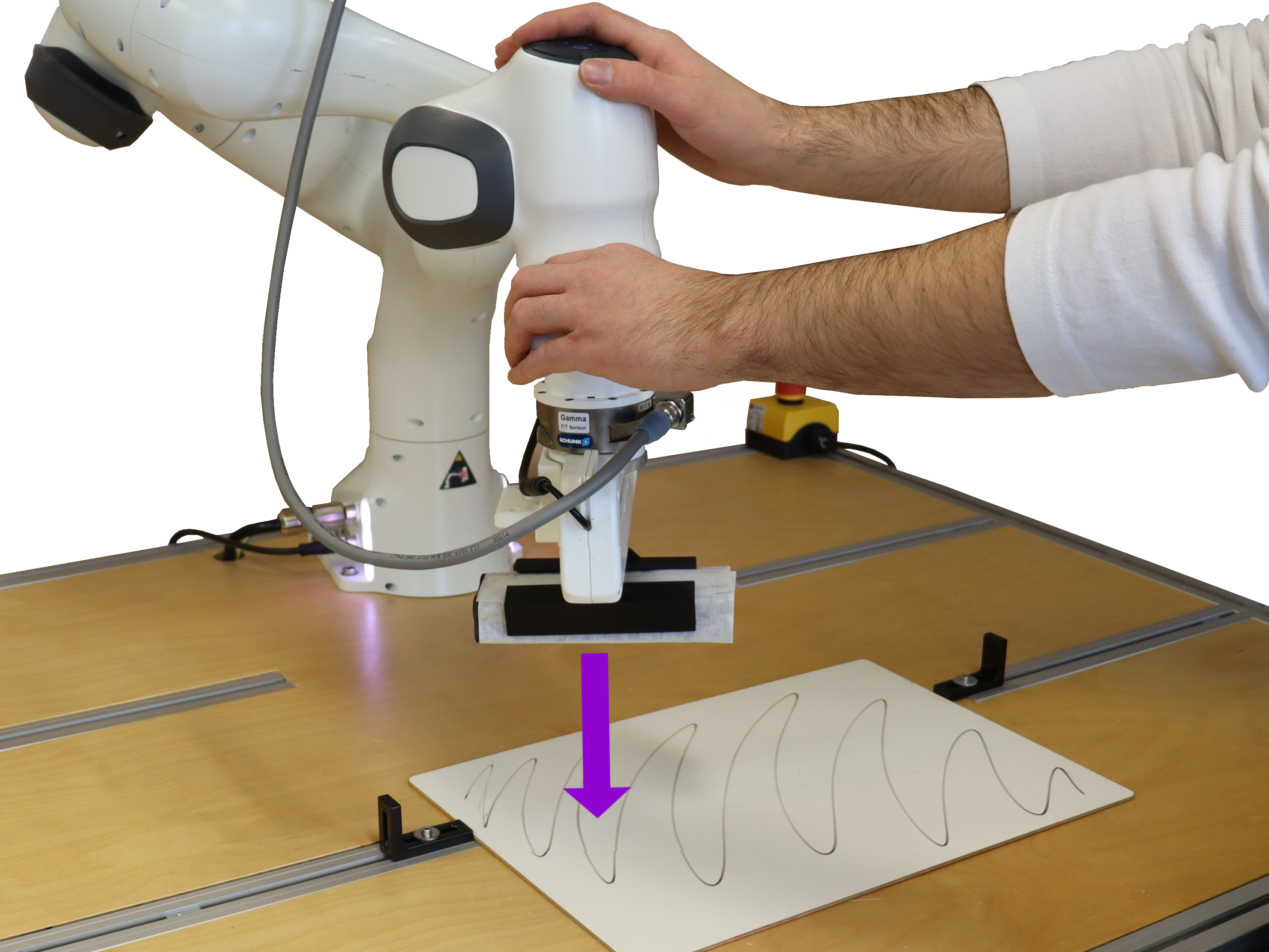}
        \caption{\textit{Guarded} motion (violet) gaining contact with the whiteboard.}
        \label{fig:guarded}
    \end{subfigure}
    \hfill
    \begin{subfigure}[t]{0.32\textwidth}
        \centering
        \includegraphics[width=\textwidth]{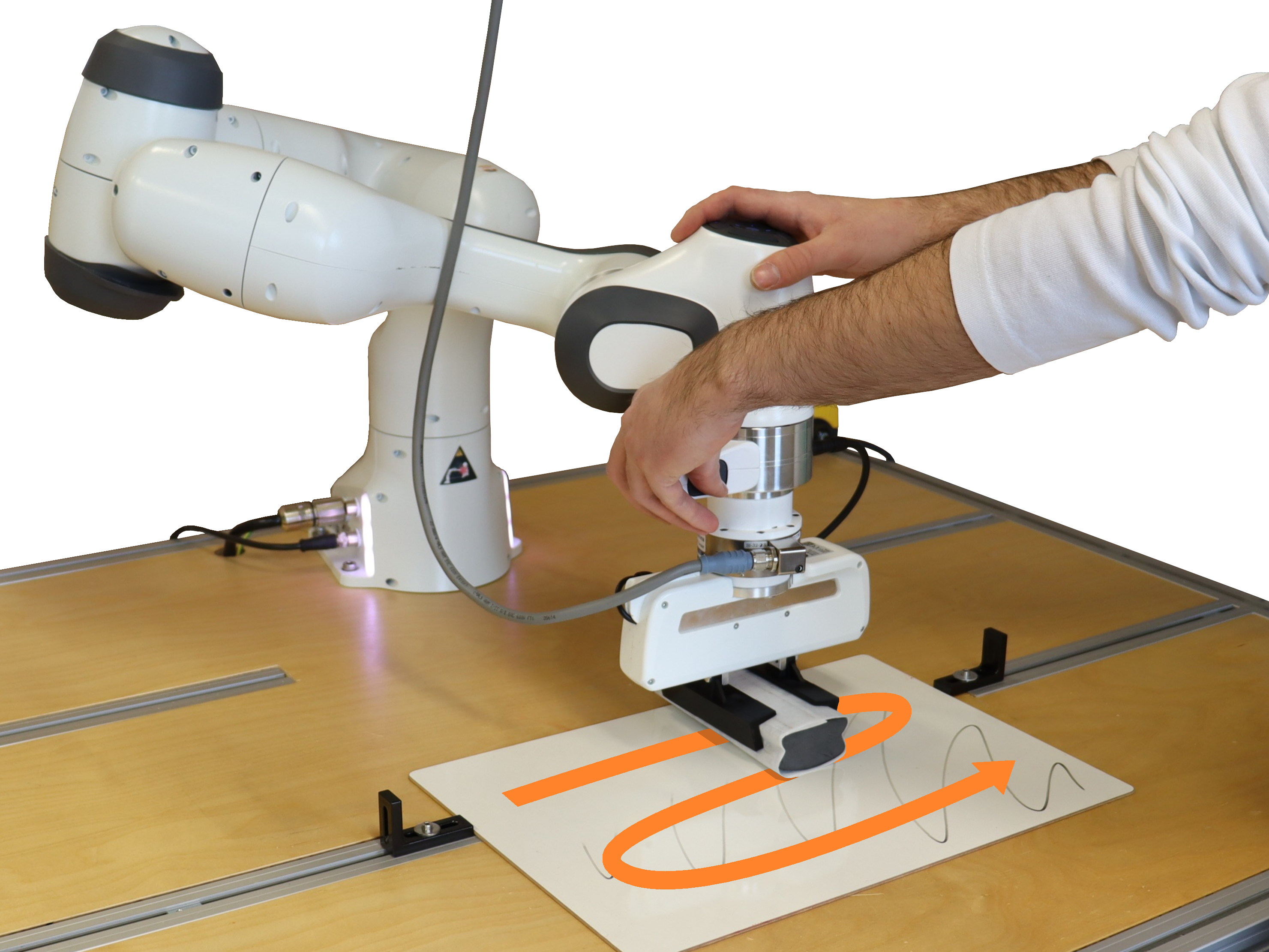}
        \caption{\textit{Compliant} motion (orange) wiping the whiteboard while applying force.}
        \label{fig:compliant}
    \end{subfigure}
    \hfill
    \begin{subfigure}[t]{0.32\textwidth}
        \centering
        \includegraphics[width=\textwidth]{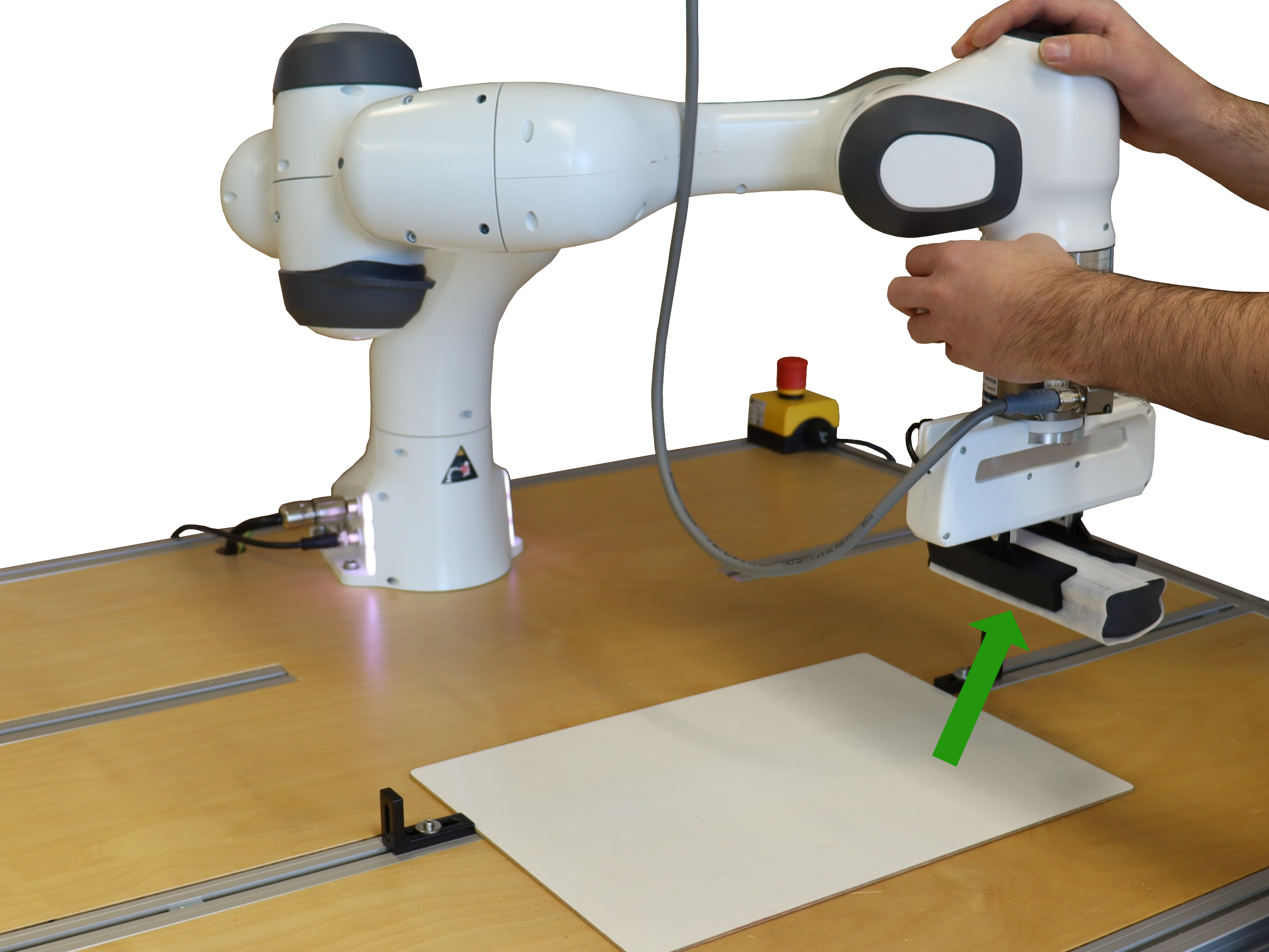}
        \caption{\textit{Free space} motion (green) away from the whiteboard.}
        \label{fig:freespace}
    \end{subfigure}
    \caption{Demonstration of a wiping task by hand guiding, containing the three described segment types.} 
    \label{fig:motion}
\end{figure*}

The user typically demonstrates a task of different motion segments during the programming process. 
For example, when wiping a whiteboard with an already gripped sponge, the segments are approaching, wiping, and departing the whiteboard (see Fig. \ref{fig:motion}). 
Segmenting and labeling the recorded path data is a lengthy and complex task that exceeds non-experts' capabilities. 
Therefore, automating this procedure is necessary for an intuitive workflow. 
The advantages of such segmentation are, for example, extraction of the end condition of the segment, optimization of the movement (acceleration, adaption through machine learning models), or visualization in the programming system.

While previous literature has proposed various labeling approaches (see Sect. \ref{sec:related_work}), we aim for a method segmenting a robot motion into universally applicable motion types related to the fundamentals of force-controlled actions \cite{Mason1981, Bruyninckx1996}.
To achieve this, we categorize motion segments into three distinct types:
(a) \textit{Guarded} motion, which terminates upon reaching a predefined force/torque condition (Fig. \ref{fig:guarded});
(b) \textit{Compliant} motion, where interaction with the environment continues until a specified trajectory is completed (Fig. \ref{fig:compliant});
(c) \textit{Free space} motion, where the robot moves without external contact (Fig. \ref{fig:freespace}).
A key advantage of such a segmentation approach is its generality, as these three motion types allow for a complete and assumption-free decomposition of any demonstrated task.

Early segmentation methods relied on rule- and threshold-based techniques \cite{Wang1996, Wang1998}, which limit adaptability across various tasks, application areas, and robotic systems. 
To overcome these limitations, we propose a research direction exploring machine learning-driven segmentation. 
By training a model to classify motion segments, we aim to eliminate the need for predefined thresholds, creating a more adaptive and parameter-free method. 
Our goal is to enable non-expert users to intuitively program general in-contact tasks, enhancing the accessibility of robot programming. 
We start by reviewing related work with a focus on machine learning approaches (Sec. \ref{sec:related_work}), followed by our proposed method (Sec. \ref{sec:method}), and conclude the extended abstract in Sec. \ref{sec:conclusion}.

\section{Related Work}
\label{sec:related_work}

Several studies have attempted to automate the segmentation and labeling of robot motion paths using machine-learning models. 
Roberge et al. \cite{Roberge2017} introduced a convolutional neural network (CNN) architecture, using force/torque data to detect insertion operations. 
Iturrate et al. \cite{Iturrate2019} extended this approach by adding translational velocity as an input, focusing specifically on insertion tasks as well. 
Their segmented sub-paths were subsequently used to generate Dynamic Movement Primitives (DMPs), highlighting the usefulness of such a segmentation. 
A relevant study by Sørensen et al. \cite{Sorensen2023} builds upon this previous work and utilizes an long short-term memory (LSTM)-CNN architecture to segment and label motion data based on end-effector velocity and applied wrenches to the environment.
Their method categorizes movements into five categories (free air, pick up, place, extraction, and insertion), solving a peg-in-hole task domain. 
Furthermore, Ji et al. \cite{Ji2022} propose a neural network (NN) utilizing LSTM and CNNs. 
Motions are classified into interaction primitives (non-contact, contact, stick, and slip). 
Here, the joint velocity and torque demonstrated through teleoperation served as input.
Next to these NN-based approaches, Fard et al. \cite{Fard2016} proposed a labeling method based on dynamic time warping combined with a k-nearest neighbor algorithm. 
This technique identifies the closest paths within a pre-labeled library of movements.
Eiband et al. \cite{Eiband2023} explored a hybrid approach, leveraging support vector machines (SVM) alongside pre- and post-conditions to classify tasks by combining machine learning with logical skill recognition.

To conclude, most recent developments favor NNs over explicit rule-based segmentation or motion library template matching.
One key reason is the ability of nNNs to learn complex relations and patterns in data \cite{Fawaz2019}. 
Furthermore, defining explicit rules for segmentation is challenging due to variations in speed, force/torque, and trajectory shape depending on the task. 
For these reasons, we also aim to adopt a NN-based approach.
As the tasks demonstrated semantics for the user is usually in cartesian space, we want to utilize this as our input data.
In contrast to the existing approaches, we want to incorporate full cartesian twists and wrenches of the end effector of the demonstrated motion.

\section{Proposed Method}
\label{sec:method}

We can formalize our wanted segmentation as follows: 
Given a time series $Z$ consisting of $T$ elements $Z_i$, with structure 
$Z_i = (v, \omega, f, \tau) \in \mathbb{R}^{12}$.
This data is recorded during hand guiding by the user in the end effector coordinate frame. 
It is given by the observed linear velocities $v \in \mathbb{R}^3$, angular velocities $\omega \in \mathbb{R}^3$, force $f \in \mathbb{R}^3$, and torque $\tau \in \mathbb{R}^3$. 
The objective of our NN is to learn a function $f: \mathbb{R}^{12 \times n} \rightarrow L$ taking $n$ elements of the time series and mapping it to one of the segment types $L = \{ \text{guarded, compliant, free space} \}$.
To enable the NN to learn this function, we want to adapt the latest LSTM-CNN network structures \cite{Sorensen2023, Ji2022} to our new input dimension and optimize them.

To generate training data for our NN, we use a 7-axis Franka Emika Panda robot arm equipped with a wrist-mounted Schunk force torque sensor to collect data from eleven different tasks (e.g., peg-in-hole, button pressing, wiping) demonstrated by different users. 
Each task consists of sequences of our motion segment types. 
We record joint and force/torque data with \SI{1}{\kilo\hertz} and compute the translational and rotational velocity of the end effector. 

Data labeling is performed manually, with the \textit{guarded} label replaced by two markers: \textit{possible start point} for the beginning of a motion segment and \textit{guarded end point} to signify a termination of a guarded segment. 
Since the exact moment a \textit{guarded} motion begins is difficult to define (i.e., distinguishing between free space and initiating a guarded approach to the surface), relying on a fixed-length segment from the preceding trajectory would be arbitrary.
Instead, we want to train our model to detect changes in user behavior at the beginning of the guarded motion.
After segmentation, we link these two additional markers together, defining a \textit{guarded} motion segment. 
Once we pair every \textit{guarded end point}, the remaining \textit{possible start points} function as a sub-segmentation for all of the resulting segments as the user changes its behavior there.
We are currently in the data collection phase and need to evaluate if the network should learn these sub-segments or directly the target function $f$.

\section{Conclusion}
\label{sec:conclusion}

Automating robot in-contact motion segmentation and labeling is a step toward enabling non-expert users to program robotic systems intuitively. 
Current segmentation methods often rely on NNs due to their ability to discern complex patterns in diverse datasets. 
Our research direction builds on this by developing an NN with an LSTM-CNN architecture using cartesian twists and wrenches as input data. 
Using these adaptions, our work aims to provide a universal approach for segmenting robot motions into task-agnostic segments.
After realizing the proposed approach, we plan to evaluate the increase of intuitiveness with and without such a segmentation in a non-expert programming framework.

\addtolength{\textheight}{-9.5cm}   

\bibliographystyle{IEEEtran}
\bibliography{IEEEabrv,references}

\end{document}